# A polarity-aware multi-relational model for signed chemical-gene interaction prediction in biological networks

Yize Zhou, Meijie Wang, Lun Yu*
Metanovas Biotech, Inc., San Jose CA 95113

**Abstract**: Predicting signed chemical-gene interactions is important for mechanistic drug discovery, network pharmacology, and drug repurposing. Existing graph-based models for biomedical networks typically predict whether an interaction exists or classify broad relation types, but they often do not explicitly distinguish activating, inhibitory, and non-polar relations within the same chemical-gene graph. Here, we propose PAMR (Polarity-Aware Multi-Relational model), a deep graph framework for predicting both polar relations (e.g., Increase and Decrease) and non-polar relations (e.g., Binding and Affect). PAMR combines graph convolutional encoding with a tensor-decomposition decoder and uses a conflict-aware Cannot-Link sampling strategy to reduce ambiguity between mutually exclusive polar relations. We further introduce two polarity-specific evaluation metrics, Polarity Discrimination Score (PDS) and CP@100, to assess whether a model separates opposite interaction types rather than only ranking edges globally. In benchmark experiments, PAMR-CL achieved a Macro AUROC of 0.9072 and a CP@100 of 0.9740, outperforming directly comparable multi-relational baselines (RGCN, GraphSAGE, TransE, and BioNet) on most global ranking metrics and on polarity-specific metrics. A nicotine case study identified two high-confidence nicotine-gene suppression predictions, S100A6 and SPP1, that were absent from the training knowledge base and supported by independent experimental literature. These results suggest that polarity-aware modeling can improve chemical-gene interaction prediction and provide a scalable computational framework for network pharmacology analysis.

## 1 Introduction

Drug repurposing has become an important strategy in pharmaceutical research and development because it can identify new therapeutic uses for existing drugs while reducing development time, cost, and risk (Pushpakom et al., 2019). For example, doxycycline has been evaluated for antiviral repurposing during the COVID-19 pandemic (Abdulaziz et al., 2022), and thalidomide, despite its historical safety concerns, has been repurposed for multiple myeloma and other cancers (Ali et al., 2012).

A key enabler of repurposing is network pharmacology, which represents interactions among drugs, targets, pathways, and diseases as biological networks (De Anda-Jáuregui et al., 2018). Network-based analysis can reveal potential new indications by connecting known drugs to disease-associated targets (Hopkins, 2008) and can also help prioritize drug combinations for complex diseases (Zimmermann et al., 2007). However, many traditional network-based methods rely on handcrafted features or shallow similarity metrics, limiting their ability to capture complex higher-order relationships in heterogeneous biomedical graphs.

*Corresponding author: lunyu@metanovas.com

Deep graph models have advanced drug repurposing and chemical-gene interaction prediction by learning representations directly from biomedical networks (Sosa et al., 2020; Zemouri et al., 2019). For example, Gaudelet et al. (2021) explored graph machine learning for drug-target interaction prediction, Zitnik et al. (2018) used graph convolutional networks to predict side effects of drug combinations, and Wang et al. (2020) applied graph convolutional methods to drug-gene interaction resources such as DGIdb (Cotto et al., 2018). Despite these advances, most models do not explicitly model the sign or polarity of chemical-gene relations (Derr et al., 2018; Du et al., 2019). In databases such as DrugBank (Wishart et al., 2018) and related chemical-gene resources, relations such as agonist, activator, and inducer imply positive effects, whereas inhibitor, blocker, and antagonist imply negative effects; other terms such as affect or binding indicate an interaction without a clear polarity. Models that ignore these distinctions may conflate activation, inhibition, and non-polar associations, reducing their mechanistic utility. Signed-network approaches have been useful in gene co-expression, protein-protein interaction, and drug-drug interaction settings (Mason et al., 2009; Yang et al., 2020; Chen et al., 2022), but polarity-aware modeling for multi-relational chemical-gene networks containing both polar and non-polar relations remains underexplored.

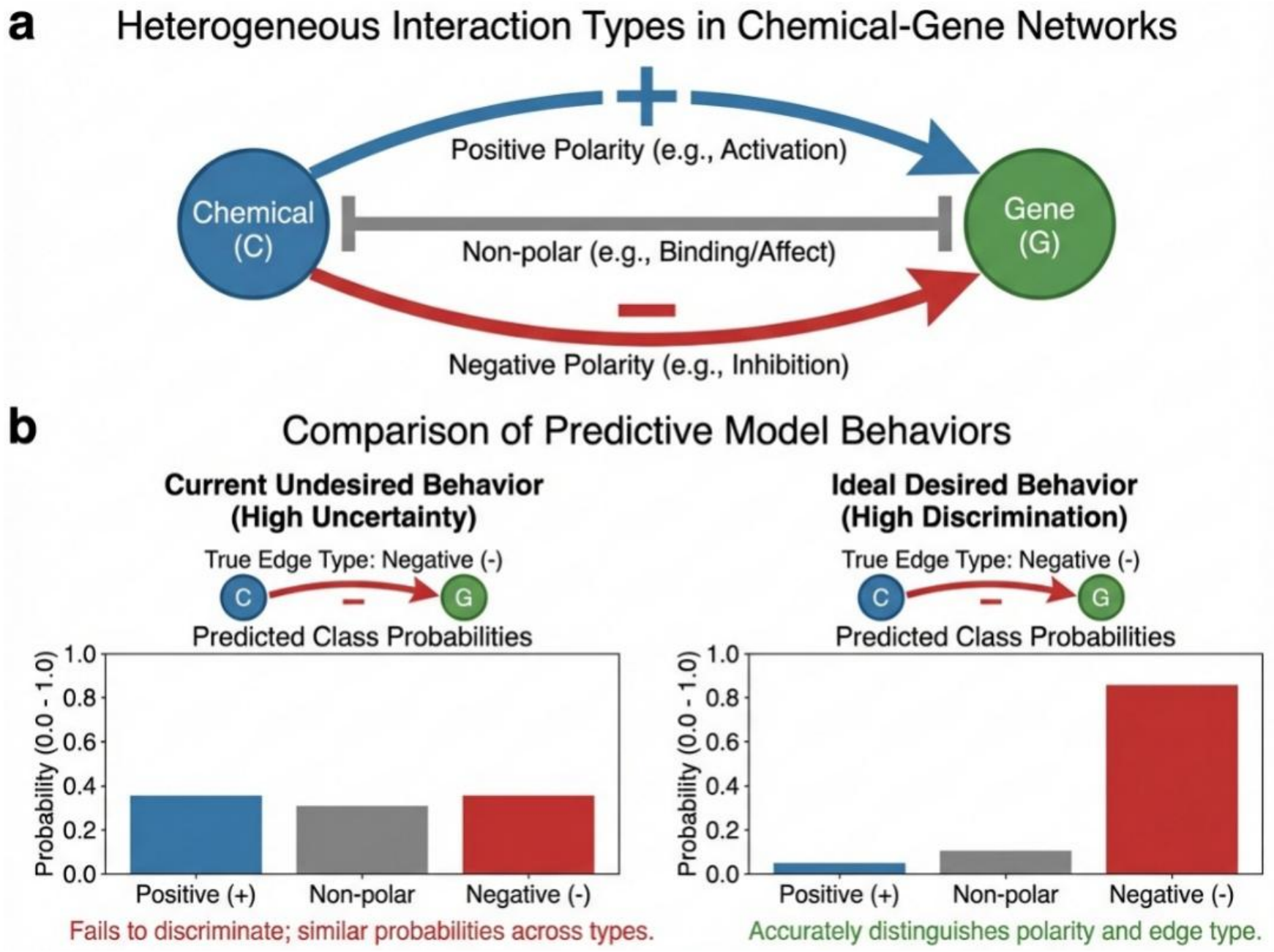

Figure 1. Chemical-gene edges can represent polar relations (e.g., positive or negative effects) or non-polar relations (e.g., binding or affect). For polar relations, the model should assign a clear probability margin between opposite interaction types rather than producing similar probabilities for activation and inhibition.

Current chemical-gene interaction prediction models generally focus on relation-type classification without explicitly training the model to separate opposite polarities. We expect a polarity-aware model to produce a larger probability margin between mutually exclusive polar relations, such as Increase and Decrease. In contrast, models that assign similar probabilities to opposite polarities fail to capture mechanistically important distinctions (Figure 1). Although signed networks provide one route for modeling polarity, conventional signed-graph formulations are not designed to simultaneously accommodate non-polar relations such as Binding and Affect. This motivates a unified framework that can model polar and non-polar chemical-gene relations in the same heterogeneous graph.

*Corresponding author: lunyu@metanovas.com

To address these challenges, we develop PAMR (Polarity-Aware Multi-Relational model), a graph learning framework for polarity-aware interaction prediction in biological networks. The main contributions of this work are summarized as follows:

- Novel polarity-aware architecture: PAMR models mutually exclusive polar relations (e.g., Increase and Decrease) while retaining non-polar relations (e.g., Binding and Affect) in a unified multi-relational chemical-gene graph.
- Conflict-aware sampling strategy: We introduce a Cannot-Link (CL) sampling strategy that discourages contradictory assignments between opposite polar relations and improves the consistency of learned representations during training.
- Polarity-specific evaluation metrics: We define Polarity Discrimination Score (PDS) and CP@100 to evaluate whether a model distinguishes opposite polarities, complementing conventional AUROC, AUPRC, and AP@20 metrics.
- Biological literature validation: In a nicotine case study, PAMR identified two high-confidence suppression predictions, S100A6 and SPP1, that were absent from the training knowledge base and supported by independent experimental literature.

# 2 Results

## 2.1 Overall prediction performance

We designed PAMR to improve chemical-gene relation prediction by integrating graph convolutional encoding with a tensor-decomposition decoder. Model performance was evaluated using Macro AUROC, Micro AUROC, Macro AUPRC, Micro AUPRC, and AP@20. Using the encoder framework of BioNet (Yang et al., 2022), we compared PAMR with BioNet and other graph representation baselines to assess the contribution of the proposed decoder and sampling strategy.

As shown in Table 1, PAMR substantially outperforms the directly comparable multi-relational baselines, including BioNet, RGCN, GraphSAGE, and TransE, across most global ranking metrics. Compared with BioNet, PAMR improves Macro AUROC from 0.7443 to 0.8790, Micro AUROC from 0.7466 to 0.8378, Macro AUPRC from 0.7176 to 0.8295, and Micro AUPRC from 0.7117 to 0.7799. These results indicate that the proposed tensor-decomposition decoder improves relation-specific representation learning in heterogeneous chemical-gene networks.

SGAN (Huang et al., 2019) reports higher scores across several metrics; however, it is included only as a contextual signed-graph reference rather than as a directly comparable baseline. SGAN uses a directed signed-network formulation in which signed edge information is explicitly available during training. PAMR instead treats polarity as a prediction target in a multi-relational heterogeneous graph that contains both polar and non-polar relations. Because these settings differ in label space, supervision signal, and graph construction, SGAN should not be interpreted as a head-to-head benchmark under identical assumptions.

*Corresponding author: lunyu@metanovas.com

GraphSAGE achieves the highest AP@20 among the directly comparable baselines, indicating stronger early retrieval for a small number of top-ranked edges. PAMR shows stronger global discrimination and calibration, as reflected by AUROC and AUPRC. Overall, PAMR provides a more balanced framework for polarity-aware chemical-gene interaction prediction under realistic multi-relational graph settings.

Table 1. Performance comparison on chemical-gene interaction prediction. RGCN, GraphSAGE, TransE, BioNet, and PAMR were evaluated under the same multi-relational heterogeneous graph setting. SGAN is reported separately as a contextual signed-graph reference because it uses signed-edge information during training and is not directly comparable to the multi-relational setting.

| **Model** | **Macro AUROC** | **Micro AUROC** | **Macro AUPRC** | **Micro AUPRC** | **AP@20** |
|---|---|---|---|---|---|
| RGCN | 0.6644 | 0.7058 | 0.6210 | 0.6633 | 0.5750 |
| GraphSAGE | 0.5515 | 0.4985 | 0.5218 | 0.4739 | **0.7523** |
| TransE | 0.6223 | 0.6665 | 0.6198 | 0.6638 | 0.7001 |
| BioNet | 0.7443 | 0.7466 | 0.7176 | 0.7117 | 0.4225 |
| PAMR | **0.8790** | **0.8378** | **0.8295** | **0.7799** | 0.6590 |
| SGAN | 0.9183 | 0.9163 | 0.9247 | 0.9342 | 1.0000 |

Note: SGAN is included as a contextual signed-graph reference rather than a directly comparable baseline because it uses directed signed-edge information during training, whereas PAMR and the other baselines treat polarity as a prediction target.

## 2.2 Effect of Cannot-Link sampling on polarity conflicts

In this study, edge conflicts refer to contradictory polarity assignments for the same chemical-gene pair, such as assigning both Increase and Decrease to a relation that should be mutually exclusive. To reduce such conflicts, the model must distinguish polar relations while preserving non-polar interaction types.

*Corresponding author: lunyu@metanovas.com

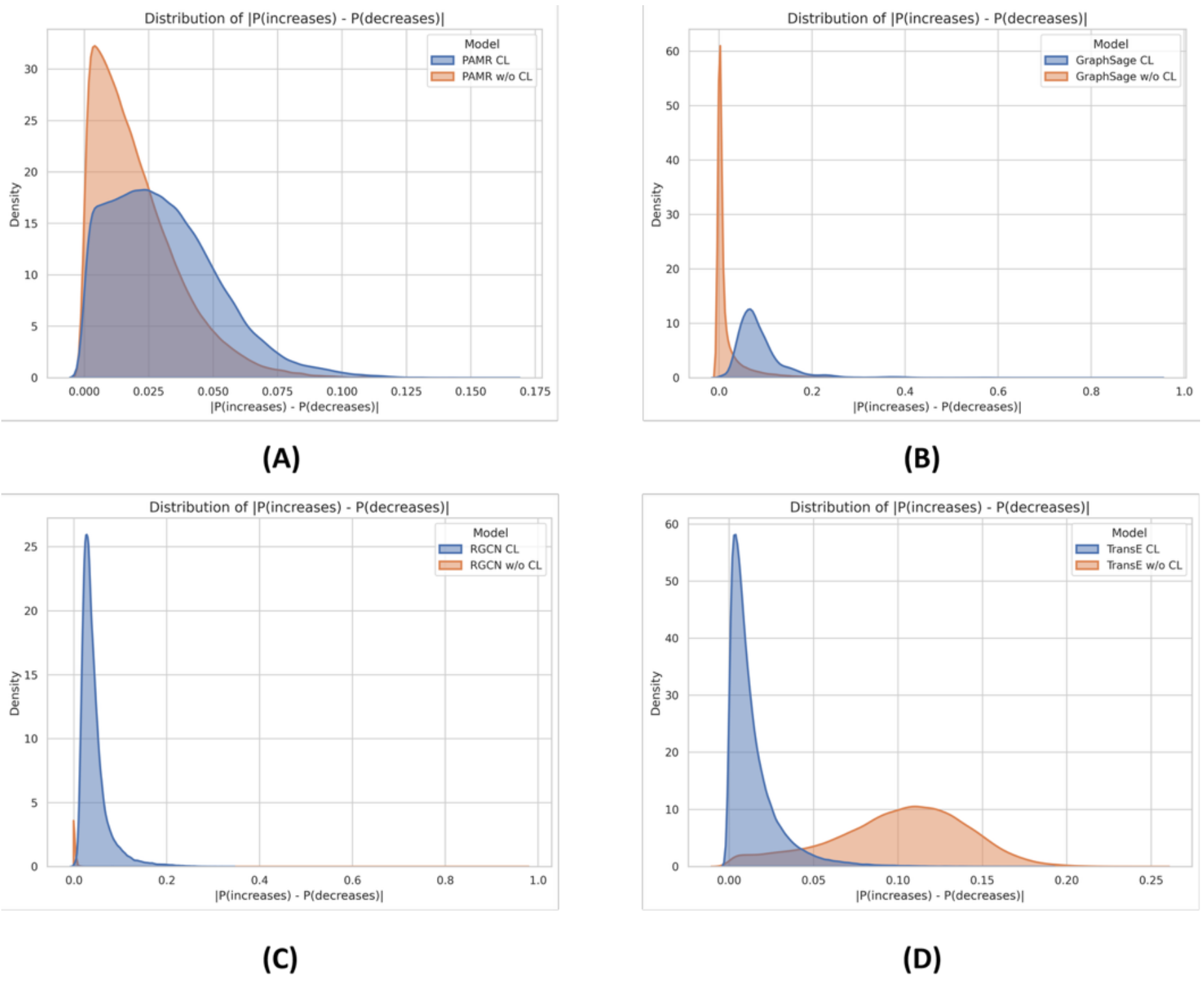

Figure 2. Density distributions of the signed polarity margin, $\Delta p$ = p(Increase) - p(Decrease), on the test set before (orange) and after (blue) applying the Cannot-Link (CL) sampling strategy. Each subplot corresponds to one model: (A) PAMR, (B) GraphSAGE, (C) RGCN, and (D) TransE. A shift away from zero and reduced overlap indicate improved discrimination between opposing polarities.

We evaluated each model's ability to distinguish polar interactions by analyzing predicted probability margins between Increase and Decrease relations on the test set (Figure 2). Greater separation between the distributions indicates stronger polarity discrimination. For PAMR (Figure 2A), CL sampling increased the separation between opposite polarities. The baseline models initially showed more overlap, indicating weaker polarity awareness. After applying CL, GraphSAGE, RGCN, and TransE also showed reduced overlap, although the separation remained less complete than in PAMR. Overall, CL sampling consistently improved the clarity of polarity prediction across models.

We further evaluated the impact of CL sampling by comparing each model with and without CL in Table 2. CL improved most AUROC and AUPRC metrics, supporting its role in improving global discrimination and calibration. PAMR-CL achieved the strongest performance among the CL-compatible multi-relational models, with Macro AUROC of 0.9072, Micro AUROC of 0.8636, Macro AUPRC of 0.8580, and Micro AUPRC of 0.7921. CL also improved several baselines, including BioNet and TransE. However, AP@20 decreased for GraphSAGE-CL and PAMR-CL, suggesting a trade-off between polarity separation and early top-k retrieval precision. Because CL regularization penalizes polarity-conflicting assignments, it can redistribute probability mass away from a small number of top-ranked edges while improving calibrated global discrimination. This trade-off should be interpreted explicitly

*Corresponding author: lunyu@metanovas.com

rather than treated as a uniform improvement across all ranking metrics.

Table 2. Performance comparison of CL-compatible multi-relational models with and without the Cannot-Link (CL) sampling strategy. The best result among these models for each metric is bolded, and improvements from CL over the corresponding base model are underlined.

| Model | Macro AUROC | Micro AUROC | Macro AUPRC | Micro AUPRC | AP@20 |
|---|---|---|---|---|---|
| RGCN | 0.6644 | 0.7058 | 0.6210 | 0.6633 | 0.5750 |
| RGCN-CL | <u>0.6744</u> | <u>0.7121</u> | <u>0.6345</u> | 0.6581 | 0.5621 |
| GraphSAGE | 0.5515 | 0.4985 | 0.5218 | 0.4739 | 0.7523 |
| GraphSAGE-CL | <u>0.7453</u> | <u>0.7151</u> | <u>0.6584</u> | <u>0.6078</u> | 0.2362 |
| TransE | 0.6223 | 0.6665 | 0.6198 | 0.6638 | 0.7001 |
| TransE-CL | <u>0.6341</u> | <u>0.6917</u> | <u>0.6269</u> | <u>0.6812</u> | 0.6434 |
| BioNet | 0.7443 | 0.7466 | 0.7176 | 0.7117 | 0.4225 |
| BioNet-CL | <u>0.7700</u> | <u>0.8045</u> | <u>0.7257</u> | <u>0.7385</u> | <u>0.4917</u> |
| PAMR | 0.8790 | 0.8378 | 0.8295 | 0.7799 | 0.6590 |
| PAMR-CL | <u>**0.9072**</u> | <u>**0.8636**</u> | <u>**0.8580**</u> | <u>**0.7921**</u> | 0.5340 |

## 2.3 Polarity-specific evaluation metrics

To further assess polarity-aware prediction, we evaluated two custom metrics: Polarity Discrimination Score (PDS), which measures discrimination between Increase and Decrease edges, and CP@100, which measures precision among the top 100 predictions ranked by polarity confidence.

As shown in Table 3, CL sampling improved polarity-specific performance for most models, particularly CP@100, indicating that CL increases confidence separation between opposite polarities. PAMR-CL achieved the best results in both metrics, with PDS of 0.6224 and CP@100 of 0.9740. Baseline models also benefited from CL, but their absolute performance remained below that of PAMR-CL.

Table 3. Performance on polarity-specific metrics, PDS and CP@100, computed on the Increase/Decrease polarity subset. The best result among CL-compatible multi-relational models is bolded, and improvements from CL over the corresponding base model are underlined.

| Model | PDS | CP@100 |
|---|---|---|
| RGCN | 0.5079 | 0.5900 |
| RGCN-CL | <u>0.5120</u> | <u>0.6700</u> |
| GraphSAGE | 0.4985 | 0.5300 |
| GraphSAGE-CL | <u>0.5068</u> | <u>0.7880</u> |
| TransE | 0.5015 | 0.4600 |
| TransE-CL | <u>0.5059</u> | <u>0.5900</u> |
| BioNet | 0.5298 | 0.7020 |

*Corresponding author: lunyu@metanovas.com

| | | |
|---|---|---|
| BioNet-CL | <u>0.5555</u> | <u>0.7660</u> |
| PAMR | 0.5999 | 0.9700 |
| PAMR-CL | **<u>0.6224</u>** | **<u>0.9740</u>** |

These results indicate that PDS and CP@100 capture model sensitivity to interaction polarity. The combination of PAMR's architecture and CL-based conflict sampling provides a stronger ability to model polar chemical-gene interactions than the evaluated baselines.

## 2.4 Retrospective nicotine case study

To demonstrate PAMR's ability to identify chemical-gene relationships that are missing from the training knowledge base, we used nicotine as a retrospective case study. Nicotine is a well-studied bioactive chemical with documented effects on gene expression, lung development, oxidative stress, and inflammatory signaling, making it suitable for literature-based validation. We examined the top 5,000 predicted nicotine interactions and applied three criteria: (1) predicted relation = Decrease; (2) the polarity margin $C = |p_{dec} - p_{inc}| \geq 0.2$; and (3) the interaction was absent from the training knowledge base. This yielded two high-confidence predictions: S100A6 and SPP1.

For both genes, PAMR assigned a markedly higher probability to Decrease than to Increase (0.7109 vs. 0.5004 for S100A6; 0.7107 vs. 0.5017 for SPP1), producing a clear suppression margin. By contrast, the BioNet baseline assigned the two polarities nearly equal probabilities (0.5029 vs. 0.4606 for S100A6; 0.5133 vs. 0.4558 for SPP1), yielding a much weaker polarity signal. Although these links were not curated in the training knowledge base, Wuenschell et al. (2004) reported that nicotine treatment significantly downregulated S100A6 and SPP1 mRNA levels in cultured embryonic mouse lung buds. S100A6 is a calcium-binding protein implicated in cellular stress responses, and SPP1 encodes osteopontin, a secreted protein involved in extracellular matrix remodeling and inflammatory signaling. This retrospective literature support suggests that PAMR can recover biologically plausible, database-missing chemical-gene suppression relations. We emphasize that this analysis provides literature-based validation rather than prospective experimental validation.

Table 4. Nicotine-gene suppression predictions absent from the training knowledge base and supported by independent literature.

| **Chem Name** | **Gene Symbol** | **Gene Name** | **Probability_decrease (PAMR / BioNet)** | **Probability_increase (PAMR / BioNet)** | **Predicted Relation** |
|---|---|---|---|---|---|
| Nicotine | S100A6 | S100 calcium binding protein A6 | 0.7109/0.5029 | 0.5004/0.4606 | decrease |
| Nicotine | SPP1 | Secreted phosphoprotein 1 | 0.7107/0.5133 | 0.5017/0.4558 | decrease |

*Corresponding author: lunyu@metanovas.com

# 3 Materials and Methods

## 3.1 Model Construction

PAMR predicts the probability of specific edge types between node pairs in a multi-relational graph. It consists of two main components: a graph convolutional network (GCN) encoder and a relational graph convolutional network with tensor decomposition (RGCNTD) decoder.

The encoder uses two stacked graph convolutional layers to extract local structural features and refine node representations through hierarchical aggregation. The decoder integrates relational graph convolutional layers with a DEDICOM tensor-decomposition module. The relational layers propagate information across relation types, while DEDICOM captures global relation-specific patterns and local entity interactions. This combination enables PAMR to predict multiple chemical-gene relation types within a unified graph framework. The relational graph convolutional encoder follows the formulation of Schlichtkrull et al. (2018), and the DEDICOM-based tensor-decomposition decoder follows Zitnik et al. (2018).

As illustrated in Figure 3, the architecture captures the multi-relational structure of chemical-gene graph data and supports relation-specific prediction for network pharmacology applications.

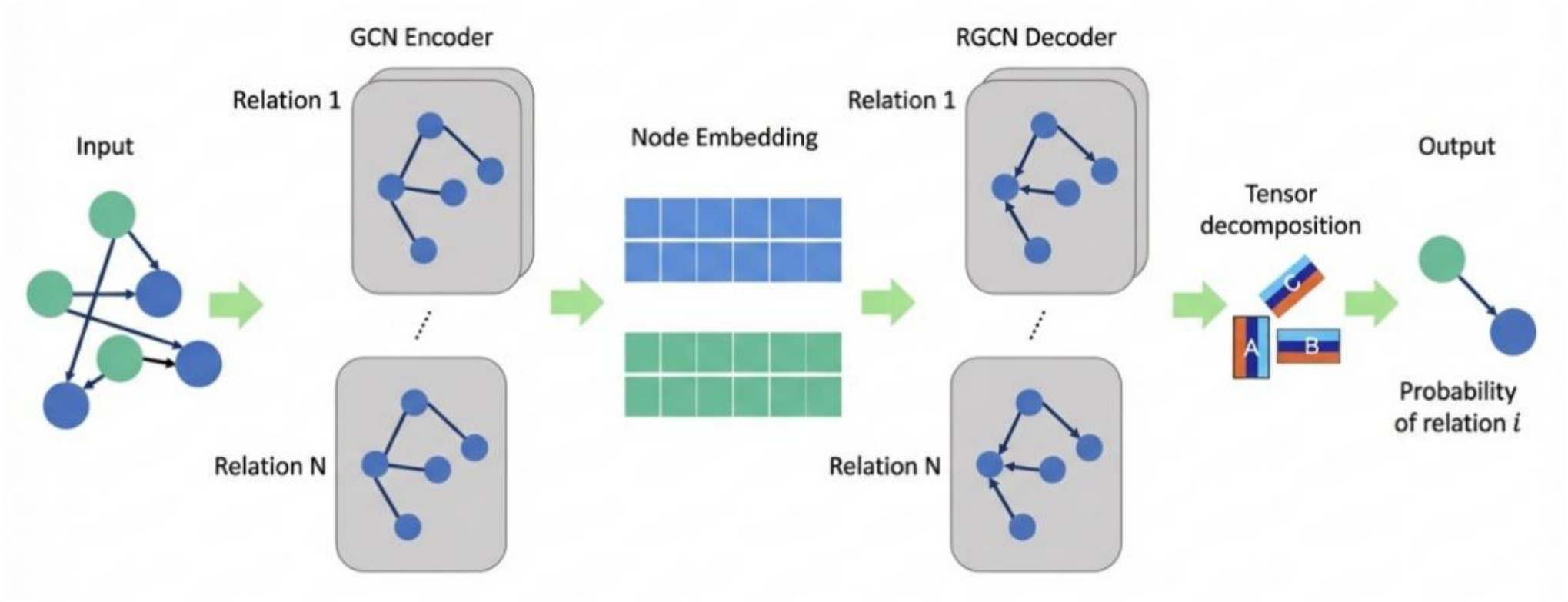

Figure 3. Overview of the PAMR model architecture.

## 3.2 Network Construction

We constructed a chemical-gene interaction network by integrating curated chemical-gene relation records and mapping chemical and gene identifiers to standardized entities. STITCH-derived chemical-protein associations (Kuhn et al., 2007) were used as chemical-gene association evidence. PubChem CIDs from STITCH were mapped to ChEBI IDs (Degtyarenko et al., 2007), and protein IDs from STITCH were mapped to NCBI Gene IDs. Chemical-gene relations were categorized into four relation types, Increase, Decrease, Binding, and Affect, according to the annotated interaction semantics in the source records. When multiple relation annotations were available for the same chemical-gene pair, each relation label was retained as a distinct multi-relational edge rather than collapsed into a single unsigned edge.

*Corresponding author: lunyu@metanovas.com

We also introduced homogeneous Chemical-Chemical and Gene-Gene interaction networks to provide additional structural information and to evaluate how auxiliary graph components affect predictive performance. The Gene-Gene interaction network was derived from 1,307,492 interaction records in the STRING database (Mering, 2003). Chemical-Chemical interaction data were obtained from 1,770,581 records in the ChEBI database. Chemical-gene relations were categorized by polarity as follows:

1. Increase: The compound exerts an activating effect on gene expression or on the activity of gene-expression products.
2. Decrease: The compound exerts an inhibitory effect on gene expression or on the activity of gene-expression products.
3. Binding: The compound binds to a gene-expression product or to a pathway associated with the gene.
4. Affect: The compound affects gene expression or the activity of gene-expression products, but the direction of the effect is unclear or not specified.

After data cleaning and removal of irrelevant molecules, the resulting heterogeneous Chemical-Gene graph contained 14,387 chemical nodes, 28,443 gene nodes, and 1,094,738 interaction records across four relation types (Table 5). We also identified 79,142 chemical-gene pairs with both polar and non-polar relationship annotations.

Table 5. Relations in the chemical-gene subgraph

| No. | Relation Name | Number of the edges | Polarity |
|---|---|---|---|
| 1 | Increase | 444,163 | Increase Polarity |
| 2 | Decrease | 434,971 | Decrease Polarity |
| 3 | Affect | 200,328 | No Polarity |
| 4 | Binding | 15,276 | No Polarity |

## 3.3 Polarity constraints for chemical-gene prediction

Chemical-gene interaction prediction must handle polar edges, where a chemical-gene pair can be associated with opposite effects such as increased versus decreased expression. We treat Increase and Decrease as mutually exclusive polar relation groups and use Must-Link (ML) and Cannot-Link (CL) constraints to guide training. ML constraints encourage relations within the same semantic group to receive high compatibility scores, whereas CL constraints penalize assignments to the opposite polarity group.

$$\begin{aligned} ML &= \left\{ \left(e_1, r_{label}, e_2\right), \left(e_1, r_{predict}, e_2\right) \middle| r_{label} \in Group_1, r_{predict} \in Group_1 \right\} \\ CL &= \left\{ \left(e_1, r_{label}, e_2\right), \left(e_1, r_{predict}, e_2\right) \middle| r_{label} \in Group_1, r_{predict} \in Group_1' \right\} \end{aligned} \quad (1)$$

*Corresponding author: lunyu@metanovas.com

Here, $r_{label}$ denotes the observed relation label and $r_{predict}$ denotes the candidate relation predicted by the model. $Group_i$ denotes a relation group, and $Group_{i'}$ denotes its opposite-polarity group. For example, if $Group_i$ contains Increase relations, then $Group_{i'}$ contains the corresponding Decrease relations. These definitions allow the model to distinguish semantically similar relations from mutually exclusive polar relations during training.

$$L = \sum_{(e_1, r_{predict}, e_2) \in ML} \left(1 - S\left(e_1, r_{predict}, e_2\right)\right)^2 + \sum_{(e_1, r_{predict}, e_2) \in CL} \left(S\left(e_1, r_{predict}, e_2\right)\right)^2 \quad (2)$$

In this loss function, $S(e_1, r_{predict}, e_2)$ denotes the decoder score for assigning relation $r_{predict}$ to the chemical-gene pair ($e_1$, $e_2$). The ML term encourages scores for same-group relations to approach 1, whereas the CL term encourages scores for opposite-polarity relations to approach 0. This design helps the model distinguish polar relations during training and improves the reliability of downstream predictions.

In our implementation, we applied the polarity constraints specifically to the Increase and Decrease relation types. Integrating this mechanism into training allows the model to learn representations that better capture mutually exclusive biological effects.

## 3.4 Baseline models and hyperparameters

We compared PAMR against BioNet, which also provides the encoder framework underlying PAMR (Section 3.5), and against three additional established graph representation learning baselines: RGCN, GraphSAGE, and TransE. RGCN uses relation-specific message passing with basis decomposition for multi-relational graphs. GraphSAGE uses neighbor sampling and inductive aggregation to preserve local structural information at scale. TransE represents entities and relations in a shared embedding space, where a head entity plus a relation vector is optimized to approximate the tail entity, providing a classical knowledge graph embedding baseline.

All baseline models were configured with two graph layers or equivalent depth, trained with the Adam optimizer, and evaluated under identical data splits and negative sampling protocols. Key hyperparameters are summarized in Table 6.

Table 6. Model parameters used for all compared models

| **Parameter** | **GraphSAGE** | **RGCN** | **TransE** | **BioNet** | **PAMR** |
|---|---|---|---|---|---|
| Hidden dimensions | [128, 128] | [100, 100] | 100 | [128, 64, 32, 16] | [128, 64, 32, 16] |
| Number of layers | 2 | 2 | 2 | 2 | 2 |
| Batch size | 8192 | 8192 | 8192 | 8192 | 4096 |
| Test batch size | 8192 | 8192 | 8192 | 8192 | 4096 |
| Learning rate | $5\times10^{-3}$ | $1\times10^{-2}$ | $1\times10^{-2}$ | $1\times10^{-3}$ | $1\times10^{-2}$ |
| Dropout | 0.2 | 0.1 | 0.1 | 0.1 | 0.1 |
| Margin (hinge loss) | 0.1 | 0.1 | 0.1 | — | — |
| Contrastive loss coefficient | — | — | — | 100 | 100 |

*Corresponding author: lunyu@metanovas.com

| Parameter | GraphSAGE | RGCN | TransE | BioNet | PAMR |
|---|---|---|---|---|---|
| Training epochs | 10 | 10 | 10 | 20 | 20 |

## 3.5 Experimental setup and leakage control

All experiments were conducted on a server equipped with an NVIDIA GeForce RTX 3090 GPU.
BioNet and PAMR were implemented within the same framework and shared identical data preprocessing, negative sampling, and evaluation protocols.

BioNet was trained with the Adam optimizer using a learning rate of 0.001, and PAMR was trained with a learning rate of 0.01. The maximum number of epochs was set to 20, with early stopping based on validation loss and a patience of 5 epochs. The best checkpoint for each model was selected using validation performance.
To reduce label leakage, train, validation, and test edges were generated after masking target relation labels from the training graph. Chemical-gene pairs with multiple relation annotations were handled at the relation level, and polarity-specific metrics were computed only on held-out Increase/Decrease edges. Negative samples were generated using the same protocol across all models so that baseline and PAMR comparisons used identical supervision and candidate sets. Fixed random seeds were used for reproducibility. Because a relation-level split may still allow the same chemical-gene pair to appear with a different relation type in another split, future evaluations should additionally report stricter pair-level or chemical-level splits.

## 3.6 Evaluation metrics

We used AUROC, AUPRC, and AP@20 (average precision among the 20 top-ranked predictions) to evaluate overall prediction performance. Macro and Micro versions of AUROC and AUPRC were calculated to account for class imbalance. In addition, we designed two polarity-specific metrics, PDS and CP@100, to evaluate whether a model distinguishes opposite polar relations rather than only ranking edges globally.

### 3.6.1 Polarity Degree for Polar Edges

We define the polarity degree C to measure how strongly a model separates the Increase and Decrease probabilities for a chemical-gene pair. Specifically, $C = |p_{inc} - p_{dec}|$, where $p_{inc}$ and $p_{dec}$ are the predicted probabilities of Increase and Decrease, respectively. Larger C values indicate greater confidence that one polarity is favored over the other, but C alone does not indicate whether the favored polarity is correct; this is evaluated by CP@k and PDS.

$$C = |p_{inc} - p_{dec}| \tag{3}$$

Because raw C values are often long-tailed, we apply a monotonic logarithmic transformation to improve visualization and analysis while preserving the ordering of predictions. This transformation compresses high-confidence values and makes differences among low-margin predictions easier to inspect.

*Corresponding author: lunyu@metanovas.com

$$C' = \frac{\ln(1+2\pi^2 C)}{\ln(1+2\pi^2)} \tag{4}$$

We observed substantial changes in the distribution of C before and after transformation. Before transformation (Figure 4A), the distribution of C was long-tailed, especially in the model without CL. A dense concentration of low C values indicated uncertainty in distinguishing Increase from Decrease interactions. After transformation (Figure 4B), the distribution showed reduced long-tail effects and greater concentration, making polarity confidence easier to visualize and compare while preserving the metric's original ordering.

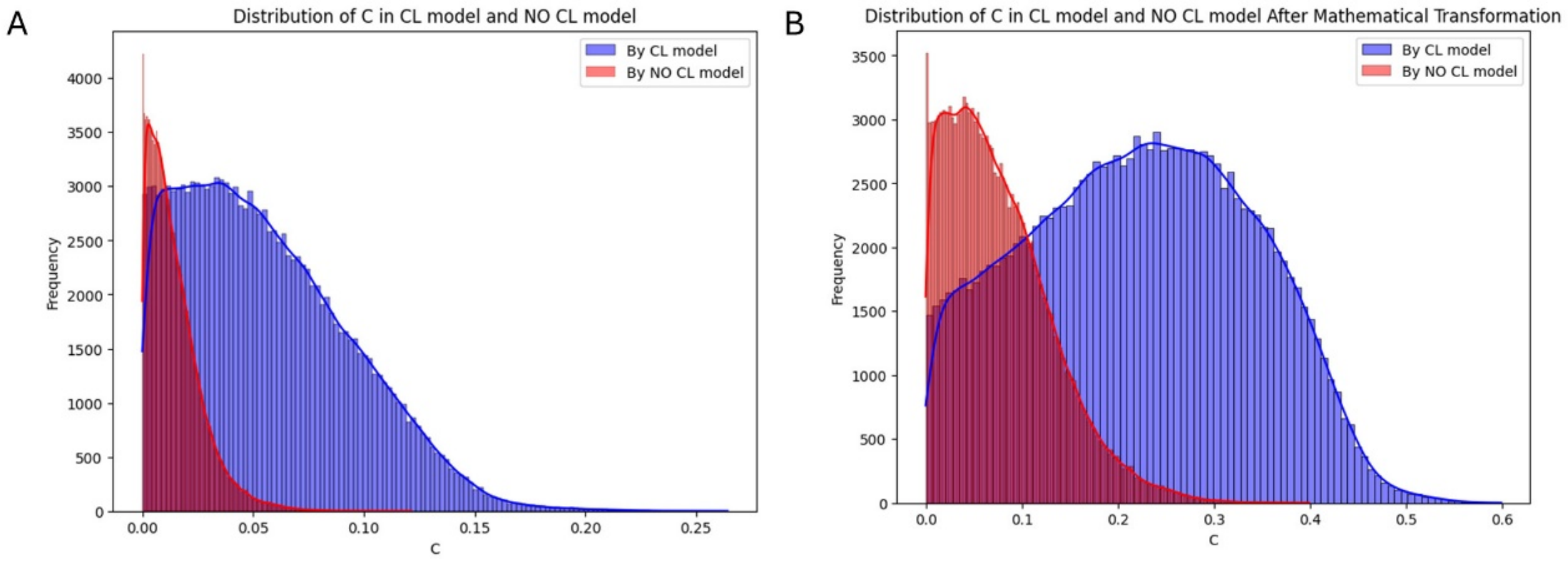

Figure 4. Distribution of the polarity degree C in models with and without CL before and after the logarithmic transformation.

### 3.6.2 C-based Precision at 100 (CP@100)

To assess model performance among high-confidence polarity predictions, we introduce C-based Precision at 100 (CP@100). This metric evaluates the accuracy of the top 100 predictions ranked by polarity degree C. The calculation involves the following steps:

1. For each instance in the dataset, compute the polarity degree C as defined in Equation (3).
2. Sort predictions in descending order by C, prioritizing predictions with the largest polarity confidence.
3. Assign the predicted polarity as the category with the larger predicted probability, and compare it with the true Increase/Decrease label. A prediction is counted as correct if the predicted polarity matches the true polarity.
4. Select the top k predictions from the sorted list and compute CP@k as shown in Equation (5):

$$CP@k = \frac{Number\ of\ correct\ predictions\ among\ the\ top\ k}{k} \tag{5}$$

CP@k evaluates precision among the most polarity-confident predictions and complements global ranking metrics such as AUROC and AUPRC.

### 3.6.3 Polarity Discrimination Score for Polar Edges

We define the Polarity Discrimination Score (PDS) to evaluate whether the model assigns a higher probability to the correct polarity for Increase/Decrease edges. For each polar test edge, the model predicts $p_{inc}$ and $p_{dec}$. A prediction is counted as correct if $p_{inc} > p_{dec}$ for an Increase edge or $p_{dec} > p_{inc}$ for a Decrease edge. PDS is then calculated as the average correctness over N polar test samples, as shown in Equation (6). Unlike C, which measures confidence

*Corresponding author: lunyu@metanovas.com

separation, PDS measures whether the direction of that separation is correct.

$$PDS = \frac{\sum \mathbb{1}(sign(p_{inc} - p_{dec}) = y)}{N} \quad (6)$$

Here, N is the total number of polar samples, and the indicator term records whether the model assigns the higher probability to the true polarity. PDS therefore reflects polarity discrimination even when the absolute probability margin is small.

# 4 Discussion

This study introduces PAMR, a polarity-aware graph model for chemical-gene interaction prediction. Prior graph neural network approaches have been effective for biomedical link prediction, but many do not explicitly distinguish activating, inhibitory, and non-polar chemical-gene relations. PAMR addresses this gap by modeling polar and non-polar interactions in a unified multi-relational framework.

A key component of PAMR is the conflict-aware CL sampling strategy, which reduces polarity ambiguity during training. CL sampling improved AUROC and AUPRC metrics and produced especially strong gains in polarity-sensitive metrics. These results support the importance of explicitly modeling polarity in biological network prediction tasks.

Comparisons with RGCN, GraphSAGE, TransE, and BioNet showed that PAMR achieved stronger global discrimination and polarity-specific performance among directly comparable multi-relational baselines. Although CL sampling improved several baseline models, its effect was most pronounced when combined with PAMR's tensor-decomposition decoder.

The comparison with SGAN highlights an important distinction in task formulation. SGAN assumes a signed-graph setting in which positive and negative edge information is available during training. By contrast, PAMR predicts relation labels in a heterogeneous graph that contains polar and non-polar relations and treats polarity as an output rather than a fully observed input feature. The SGAN results therefore provide contextual reference rather than a strict head-to-head benchmark. Future comparisons should include signed-graph baselines adapted to the same multi-relational label space to further isolate architectural effects from task-formulation differences.

Signed-graph attention architectures may require separate treatment of positive and negative edges, which can complicate direct extension to heterogeneous graphs that also contain non-polar relations. PAMR combines relation-aware message passing with tensor decomposition, providing a compact alternative for multi-relational chemical-gene prediction.

These findings suggest that polarity-aware modeling should be evaluated under realistic data assumptions, where polarity labels may be incomplete and many chemical-gene pairs remain unannotated. Future work can extend PAMR by incorporating additional biological data types, temporal or dose-dependent effects, and interpretability methods for downstream biological and clinical applications.

*Corresponding author: lunyu@metanovas.com

Several limitations should be acknowledged. First, chemical-gene interactions were aggregated from STITCH and ChEBI and may inherit curation bias, missing annotations, and noisy polarity labels, especially for less-studied compounds. Second, the current evaluation primarily uses a relation-level split; although target relation labels are masked, stricter pair-level or chemical-level splits would provide a more conservative assessment of generalization. Third, the nicotine case study provides retrospective literature support for two predictions, but broader prospective validation across additional chemicals and gene families would strengthen biological generalizability. Fourth, polarity was evaluated only for the Increase/Decrease relation pair; extension to other polar relation pairs, such as agonist/antagonist or promoter/repressor, remains to be tested. Finally, the current model does not explicitly capture context- or dose-dependent polarity reversal, which is biologically important and should be addressed in future work.

# 5 Conclusion

We presented PAMR, a polarity-aware deep graph model for chemical-gene interaction prediction. PAMR models polar and non-polar relations within a unified multi-relational framework and combines graph convolutional encoding, tensor-decomposition decoding, and conflict-aware sampling.

PAMR-CL achieved the strongest AUROC/AUPRC and polarity-specific discrimination among directly comparable multi-relational baselines, while SGAN remained a contextual reference under a different signed-graph formulation. These results demonstrate the value of explicitly modeling polarity in biological networks and provide a practical framework for polarity-aware chemical-gene interaction prediction.

**Declaration of Competing Interest**
The authors declare that they have no known competing financial interests or personal relationships that could have appeared to influence the work reported in this paper.

*Corresponding author: lunyu@metanovas.com

*Corresponding author: lunyu@metanovas.com

*Corresponding author: lunyu@metanovas.com